\documentclass{commonstack}

\usepackage[utf8]{inputenc}
\usepackage[T1]{fontenc}
\usepackage{array}
\usepackage{makecell}
\usepackage{textgreek}
\usepackage{subcaption}
\usepackage{amsmath}
\usepackage{amssymb}
\usepackage{mathtools}
\usepackage{amsthm}
\usepackage{colortbl}
\usepackage{wrapfig}
\usepackage{mathpazo}
\usepackage{float}
\usepackage{algorithm}
\usepackage{algorithmic}
\usepackage{listings}
\usepackage{multirow}
\usepackage{siunitx}
\usepackage{tikz}
\usepackage{enumitem}
\usepackage{pifont}
\usepackage{tabularx}
\usepackage{adjustbox}
\usepackage{xurl}

\title{TwinRouterBench: Fast Static and Live Dynamic Evaluation for Realistic Agentic LLM Routing}

\author{%
Pei Yang\textsuperscript{1,*} \quad
Wanyi Chen\textsuperscript{2,*} \quad
Tongyun Yang\textsuperscript{3} \quad
Pengbin Feng\textsuperscript{4} \quad
Jiarong Xing\textsuperscript{5} \quad
Wentao Guo\textsuperscript{1} \\
Yuhang Yao\textsuperscript{6} \quad
Yuhang Han\textsuperscript{7} \quad
Hanchen Li\textsuperscript{8} \quad
Xu Wang\textsuperscript{9} \quad
Yuan Gao\textsuperscript{1} \quad
Zeyu Wang\textsuperscript{10} \quad
Jie Xiao\textsuperscript{1} \\
Anjie Yang\textsuperscript{1} \quad
Liang Tian\textsuperscript{1} \quad
Lynn Ai\textsuperscript{1} \quad
Eric Yang\textsuperscript{1} \quad
Tianyu Shi\textsuperscript{1,\ensuremath{\dagger}}%
}
\affiliation{%
\textsuperscript{1}Gradient \quad
\textsuperscript{2}Soochow University \quad
\textsuperscript{3}Independent Researcher \quad
\textsuperscript{4}Amazon \quad
\textsuperscript{5}Rice University \\
\textsuperscript{6}Beijing Institute of Technology \quad
\textsuperscript{7}Carnegie Mellon University \quad
\textsuperscript{8}The Hong Kong University of Science and Technology \\
\textsuperscript{9}University of California, Berkeley \quad
\textsuperscript{10}University of the Chinese Academy of Sciences \\
\textsuperscript{*}Equal contribution \quad
\textsuperscript{\ensuremath{\dagger}}Corresponding author%
}

\date{May 8, 2026}
\papertype{arXiv preprint}
\correspondence{tianyu@gradient.network}
\sourcecode{https://github.com/CommonstackAI/TwinRouterBench}

\abstract{LLM routing matters most in long-horizon applications such as coding agents,
deep research systems, and computer-use agents, where a single user request
triggers many model calls. Routing each call to the cheapest sufficient model can cut
costs without sacrificing quality, yet existing router benchmarks evaluate
routers only on one-shot prompts. They never expose the router-visible
prefix at an intermediate agent step, never test whether a cheaper replacement
preserves downstream task success, and often rely on online LLM judges at
evaluation time.
We introduce \textbf{TwinRouterBench}, a step-level routing benchmark with two
tracks. The \emph{static track} provides 970 router-visible prefixes from 520
instances across SWE-bench, BFCL, mtRAG, QMSum, and PinchBench, each paired
with an execution-verified target tier estimated under a released
downgrade-and-cascade protocol; scoring is deterministic arithmetic
over tier labels, trajectory membership, and token costs, with no online
evaluator-side LLM judge.
The \emph{dynamic track} supplies a harness that supports the full
500-case SWE-bench Verified suite. This paper and the public leaderboard
report one locked list of 100 instances: 18 Astropy and 82 Django, disjoint
from the static SWE supervision split, and not a random sample of the
500-case suite. At each
LLM call the router selects a concrete model from a locked pool, and success
is measured by official task resolution and realized API spend. The two
tracks support fast offline iteration followed by end-to-end validation
under live agent execution. Code and data are available at
 \url{https://github.com/CommonstackAI/TwinRouterBench}.
}

\begin{document}

\maketitle

\section{Introduction}
\label{sec:intro}

\begin{figure}[t]
\centering
\includegraphics[width=\linewidth]{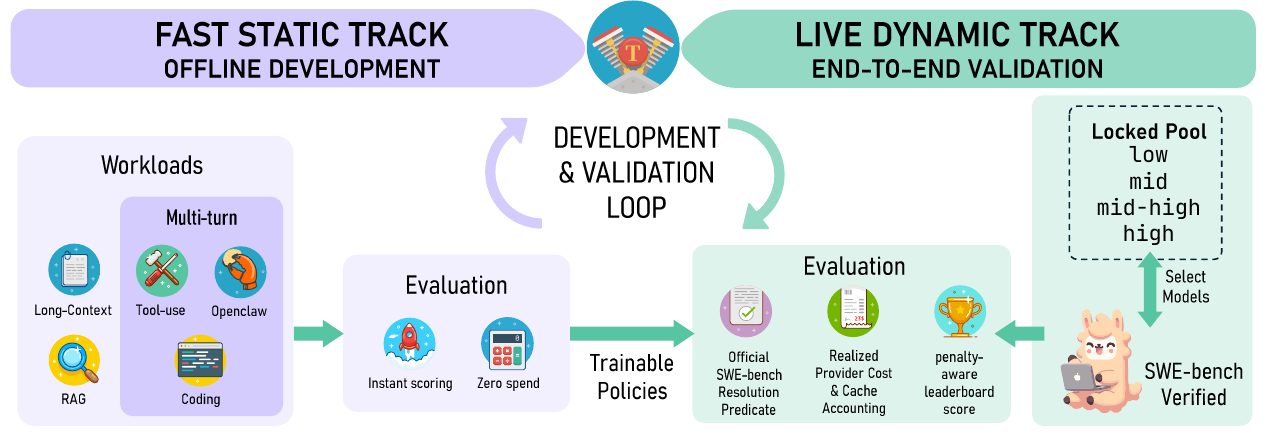}
\caption{Overview of TwinRouterBench. The benchmark provides a fast
static track for offline router development and a live dynamic track
for end-to-end validation. The static track covers 970 step-level rows
from 520 instances across five workloads, each with an
execution-verified target tier; the dynamic track runs routers on
SWE-bench Verified with realized API cost.}
\label{fig:overview}
\end{figure}

LLM routers select which model should handle each
call~\citep{feng2024graphrouter,stripelis2024tensoropera}. In production
multi-turn agents, each call carries a rich prefix---chat history, retrieved
passages, shell logs, tool outputs, and partial code edits---and a cheaper
choice at step $i$ may only fail downstream. Existing benchmarks evaluate
routing on isolated one-shot prompts; they do not expose these prefixes or
test whether a per-step decision breaks a later step.

The expensive unit in modern applications is rarely a one-shot query. Coding
agents solve GitHub issues through many rounds of exploration, patching, and
verification \citep{jimenez2023swebench}; tool systems maintain state over
serial and parallel calls \citep{patil2025bfcl}; retrieval and summarization
workloads carry long multi-turn context
\citep{katsis2025mtrag,zhong2021qmsum}. A cheap model may suffice for a file
lookup, while a later patch-writing step may require the strongest tier. A
useful step-level benchmark must cover these workloads, ground labels in
execution outcomes rather than model-based judgement alone, and support both
cheap deterministic offline scoring and end-to-end execution with realized
API cost as the definitive validation. We therefore maintain two distinct
tracks.

RouterBench, LLMRouterBench, and RouterArena evaluate query-level routing on
complete one-shot prompts
\citep{hu2024routerbench,li2026llmrouterbench,lu2025routerarena}. Such a
router can still be applied mid-trajectory by feeding the full conversation
history as a single query, but its training and evaluation data are one-shot
prompts, so neither tells us whether a cheap choice at step $i$ breaks a
later step. We expose the full prefix a router observes before each LLM call (chat
history, retrieved passages, tool outputs, shell traces, and code diffs)
as a first-class input, and we verify each label by running the task to
completion: if the trajectory still passes without increasing its number of
LLM-call steps after downgrading, we infer the replaced step was sufficient
under the construction protocol.
TRIM studies reasoning-step escalation in math
\citep{kapoor2026trim}; our setting is agentic and multi-turn.

Building such a benchmark faces two main challenges.
First, using a frontier model as the router is unreliable: in our
Claude Opus 4.6 diagnostic, a single-template LLM router predicts
\texttt{high} for only 7 of 147 verified-\texttt{high} SWE-bench steps and
fails every SWE trajectory.
Second, multi-turn trajectories create combinatorial label complexity: with
$n$ candidate models and $m$ steps, exhaustive verification requires $n^{m}$
runs even when the trajectory length is fixed, because the prefix at step
$i{+}1$ depends on the model chosen at step~$i$. Practical label
construction must therefore approximate this space with
affordable protocols.

We introduce \textbf{TwinRouterBench}, a step-level routing benchmark with a
fast static track for offline development and a live dynamic track for
end-to-end validation.

\paragraph{Contributions.}
\begin{itemize}[leftmargin=*, itemsep=1pt, topsep=2pt]
\item \textbf{A fast static track for realistic routing supervision.}
The static track covers the workloads where step-level routing pays off
most (long-context, multi-turn, tool-use, RAG, summarization, and
code-repair) and exposes the exact prefix a router observes before each LLM
call. Its
970 labels are execution-verified estimates under a fixed
downgrade-and-cascade protocol with tier-pool cascade verification,
mixed-prefix reconstruction, and manual audit on multi-step executable
workloads. Scoring is deterministic arithmetic over labels, trajectory
membership, and token costs: evaluation takes milliseconds and requires no
online evaluator-side LLM judge. The same data also supports training; a
logistic router trained on the static labels achieves the best cost--success
trade-off among the policies we evaluate.
\item \textbf{A live dynamic track for software-agent routing.}
The dynamic harness supports the full 500-case SWE-bench Verified suite.
This paper and the public leaderboard report one locked list of 100
instances: 18 Astropy and 82 Django, disjoint from the static SWE supervision
split, and not a random sample of the 500-case suite. At each LLM request
the router selects a concrete model
from a locked pool; scoring uses the official SWE-bench resolution predicate,
realized provider spend, cache accounting, and a flat unresolved-instance
penalty in the leaderboard bill (\S\ref{sec:dynamic-scoring}), with no
evaluator-side LLM judge.
\item \textbf{A two-track development and validation loop.}
The static track provides cheap, trainable supervision for rapid iteration.
The dynamic track tests whether the resulting policy survives live execution.
We report them side by side so that offline and end-to-end evidence remain
distinguishable.
\end{itemize}

\section{Related Work}
\label{sec:related}

\paragraph{Query-level routing and cascades.}
An early line of work treats model selection as an inference-time decision over
quality and cost for a whole query. FrugalGPT learns a cascade over heterogeneous APIs
to reduce cost \citep{chen2024frugalgpt}. AutoMix queries a smaller model
first, estimates reliability via self-verification, and escalates when needed
\citep{aggarwal2023automix}. Hybrid LLM routes between a local and a cloud
model based on predicted difficulty \citep{ding2024hybrid}. RouteLLM trains
routers from preference data and studies cross-pair transfer
\citep{ong2024routellm}. A recent survey formalizes the shared objective as
selecting a model subject to a cost budget \citep{varangotreille2025doing}.
These methods effectively abstract model selection on complete queries, but
they leave the intermediate calls inside multi-turn agent trajectories
under-explored, even though tool outputs and partial state already shape the
prefix.

\paragraph{Router benchmarks.}
RouterBench collects over 405k inference outcomes across diverse single-prompt
tasks, providing the first large-scale testbed for comparing routing algorithms
\citep{hu2024routerbench}. LLMRouterBench scales to 400k+ instances,
21 datasets, and 33 models under unified performance and cost metrics
\citep{li2026llmrouterbench}; although it includes 500 SWE-Bench issues,
SWE-Bench is configured as a single-query $0$-shot \textsc{Pass@1} task with
no agent scaffold and no intermediate routable call. RouterArena adds an open
comparison platform with difficulty levels and automated leaderboards
\citep{lu2025routerarena}. Triage makes one tier decision per SWE-bench Lite
issue using code-health features \citep{madeyski2026triage}. These benchmarks are valuable for comparing query-level routers, but even
when multi-step tasks such as SWE-Bench are included, routing reduces to
picking one model per issue from a pre-computed outcome matrix; the
intermediate retrieval, analysis, and debugging calls within an issue cannot
be routed individually to the cheapest sufficient model at each step.

\paragraph{Step-level routing.}
TRIM studies stepwise routing for multi-step math reasoning by identifying
critical reasoning steps and escalating them to larger models, and shows
that selective per-step upgrading can preserve accuracy while reducing
average cost \citep{kapoor2026trim}. Math reasoning, however, is a
relatively self-contained setting that lacks tool outputs, retrieval
contexts, shell traces, and code state, all of which are central to real
agent deployments. TRIM also derives labels from hypothetical reasoning
traces without executing the downgraded trajectory end-to-end to verify
task success. For executable long-context agent workloads the core
labeling problem persists: given the current prefix, which model tier is
cheap enough yet still preserves final task resolution?

\section{Problem Formulation and Benchmark Overview}
\label{sec:benchmark-overview}

\subsection{Step-level routing}

A multi-turn LLM agent produces a trajectory $\tau = (s_1, s_2, \ldots, s_N)$ of LLM calls. At step $i$, the agent holds a message history $M_i = [m_1, \ldots, m_k]$ (a \emph{prefix}: system prompt, user instruction, previous assistant messages, tool outputs, retrieval chunks), emits a new assistant message $a_i = \mathcal{R}(M_i)$, executes any tool calls in $a_i$ against an external environment to produce observation $o_i$, and appends $a_i, o_i$ to the history: $M_{i+1} = M_i \parallel [a_i, o_i]$. The agent halts when it submits or when a budget is exhausted.

\subsection{Conditional per-call routing}
\label{sec:conditional-per-call-routing}

A standard LLM router maps a query to a candidate model to optimize quality and
cost; recent surveys formalize this as selecting from a model set subject to a
cost budget \citep{varangotreille2025doing}. TwinRouterBench studies the
conditional per-call version: the query is the full router-visible prefix $x_i$
before the next LLM call, including system messages, user requests, prior
assistant messages, tool outputs, retrieval snippets, logs, and partial code
edits. A step-level router is a conditional decision rule
\[
\pi : x_i \mapsto t_i \in \mathcal T,
\]
where
\[
\mathcal T =
\{\texttt{low},\texttt{mid},\texttt{mid\_high},\texttt{high}\}
\]
is an ordered set of cost and capability tiers. Each tier maps to a pool of models
with similar capability and cost; the concrete model is decided downstream by a
tier-to-model mapping, keeping public rows free of vendor model IDs.

The ideal conditional target tier is the cheapest tier that can
correctly handle the current step given the prefix $x_i$.
Let $\mathcal M_t\subseteq\mathcal M$ denote the model pool for tier
$t$, and let $V_i(m;x_i)=1$ indicate that model $m$ produces a sufficient
response for the current step under prefix $x_i$. For one-shot workloads,
$V_i$ is the task-level success predicate. For multi-turn agentic workloads,
$V_i(m;x_i)=1$ means that model $m$ correctly solves the problem posed at
step $i$; when each step is solved correctly, the trajectory as a whole is
more likely to succeed. The ideal target is
\[
t_i^\star
=
\min\left\{
t\in\mathcal T:
\exists m\in\mathcal M_t
\ \text{such that}\
V_i(m;x_i)=1
\right\}.
\]
$t_i^\star$ is a conditional per-call quantity, not a globally optimal joint
policy over the full trajectory. At deployment time a router observes only the
current prefix $x_i$: it cannot change earlier outputs and cannot choose future
calls before their prefixes exist.

Because per-step correctness cannot always be judged in isolation, we
approximate $V_i$ through end-to-end execution: a downgraded step is
accepted when the full trajectory still passes with the same number of
steps, which lets us infer that the replaced step was handled correctly.
The released label $\hat{t}_i$ is therefore an execution-verified
estimate of $t_i^\star$ under our fixed downgrade-and-cascade protocol
(\S\ref{sec:dataset-construction}), supplemented by a manual audit
that cross-checks this inference on a random sample of steps
(\S\ref{sec:dataset-construction}, Manual audit). A tier $t$ is accepted
when at least one model in $\mathcal M_t$ resolves the mixed-model
trajectory. We use a fixed 11-model pool spanning four tiers; the full
list and grouping are in the appendix
(Table~\ref{tab:appendix-model-pool}). Changing the pool, tier
membership, harness, or verification protocol changes $\hat{t}_i$ and
constitutes a new benchmark version.

\subsection{Corpus}

The release contains 970 step-level rows from 520 trajectory instances across
five benchmarks; per-benchmark instance and step counts (and prefix-token
medians) are summarized in Appendix Table~\ref{tab:corpus-overview}. A trajectory is one
successful end-to-end run of a strong model on the original task; each step
corresponds to one LLM call inside that trajectory, and its prefix is exactly
the messages the router would see before that call
(\S\ref{sec:dataset-construction} describes the full construction). Every
row has the shape \texttt{id}, \texttt{benchmark}, \texttt{instance\_id},
\texttt{step\_index}, \texttt{total\_steps}, \texttt{messages},
\texttt{target\_tier}, and \texttt{target\_tier\_id}. No vendor model IDs
appear in the public JSONL, only the tier label. Full row examples are in
the appendix.

Appendix Table~\ref{tab:tier-distribution} summarizes the distribution of cost-saving
opportunities across tiers. The \texttt{low} tier tests whether a router can
avoid unnecessary expensive calls while preserving task success. The two middle
tiers are transitional capability bands; routers with three output classes can
merge them into a single middle band or map through the released tier ids.
SWE-bench contributes most \texttt{high}-tier rows, signaling when a router
must stay conservative in difficult code-repair steps.

mtRAG and QMSum contribute single-call rows whose prefixes may contain
multi-turn context. BFCL contains both single-turn and multi-turn function
calling: 130 instances yield 248 rows, with 29 instances having more than one
routed step. SWE-bench and PinchBench contribute multi-step agent traces. We
include all cases because production routers encounter both single-call
long-context states and sequential agent states.

\subsection{Static evaluation without an online judge}

Static scoring is deterministic arithmetic over tier labels, trajectory
membership, and token costs; the four metrics \textsc{RowPass},
\textsc{RowExact}, \textsc{TrajPass}, and \textsc{CostSave} are defined in
\S\ref{sec:static-scoring}.

\subsection{Dynamic SWE-bench track}

The dynamic track provides a harness that supports the full 500-case
SWE-bench Verified suite. This paper and the public leaderboard report one
locked list of 100 instances: 18 Astropy and 82 Django, disjoint from the
static SWE supervision split, and not a random sample of the 500-case suite.
At each LLM call the router
chooses a concrete model from the locked pool given the current prefix,
available models, cache state, and budget. The track reports three results:
resolve rate (the official SWE-bench predicate), realized API spend, and a
leaderboard bill that adds the same flat unresolved-instance penalty on top
of API cost for every policy (\S\ref{sec:dynamic-scoring}).

\noindent\begin{minipage}{\linewidth}
We also implement an OpenAI-compatible routing interface for OpenCode,
allowing a routing policy to select a model at each call using the current
conversation and tool definitions (Appendix~\ref{app:artifact}).\par
\end{minipage}\par

\section{Dataset Construction}
\label{sec:dataset-construction}

\begin{figure}[t]
\centering
\includegraphics[width=\linewidth]{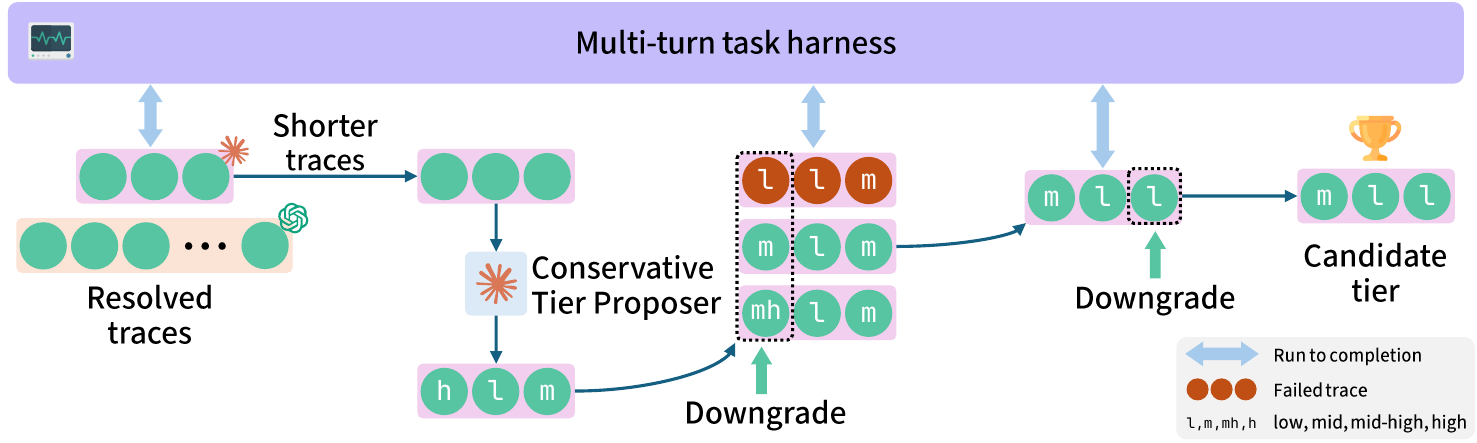}
\caption{TwinRouterBench construction pipeline. For each multi-turn case, the
pipeline starts from a successful strong-model trajectory and progressively
downgrades individual steps to cheaper tiers via execution-verified search,
producing the verified tier label for every LLM call in the trace.}
\label{fig:pipeline}
\end{figure}

Figure~\ref{fig:pipeline} illustrates how each multi-turn case is progressively
downgraded from a successful strong-model trajectory to produce per-step
verified tier labels through execution-verified search.

\paragraph{Seed from successful strong trajectories.}
We first collect trajectories that resolve under a strong-model setting. For
SWE-bench, Claude Opus 4.6 or GPT-5.4 running through
\texttt{mini-swe-agent} must submit a patch that passes the official
\texttt{FAIL\_TO\_PASS} tests; BFCL, mtRAG, QMSum, and PinchBench~\citep{pinchbench} use their
respective harness pass predicates. Failed seeds are dropped so that routing
labels do not conflate task difficulty with routing error.

\paragraph{Search lower tiers under causal prefixes.}
For each surviving trajectory, Claude Opus 4.6 provides a search hint:
whether a step appears downgradeable and, if so, the most aggressive lower tier
$h_i$ to probe first. The hint is a pruning device, never a label. We then run
a greedy sequential-locking downgrade search (Appendix Algorithm~\ref{alg:seq-lock}).
At step $i$, previously locked steps keep their verified tiers, future steps
use the strong tier, and candidates for step $i$ are tried from cheapest to
strongest starting at $h_i$. Because per-step correctness cannot always be
observed directly, we accept a candidate when the full trajectory still passes
and its number of LLM-call steps does not exceed that of the current accepted
trajectory. A shorter successful continuation is allowed. If accepted, it
becomes the current trajectory for subsequent steps; if no candidate passes,
the step stays at the strong tier. This
reduces the naive $|T|^N$ grid to $\mathcal{O}(|T|N)$ trials while ensuring
later prefixes contain actual degraded outputs from earlier routed steps.
Locked prefixes are replayed from a response cache keyed by instance, step,
model, and generation parameters; cache entries are invalidated when an
upstream lock changes.

\paragraph{Verify tiers as pools, not single models.}
A tier label should mean that the tier pool can solve the step, not that one
model happened to pass. We run a three-model cascade for non-\texttt{low}
labels: a tier is accepted if any of the three pool models passes under the
mixed-model trace. In a SWE-bench last-step audit, the cascade recovered 28 of
33 downgradeable labels versus 15 of 33 under single-model probing, confirming
that single-representative search is too brittle for supervision.

\paragraph{Harden open-ended pass predicates.}
Some benchmarks (mtRAG, QMSum) rely on LLM-as-judge evaluation whose default
prompts and rubrics are not strict enough. During early development we manually
inspected a random sample of judge verdicts and found pseudo-passes where
surface-level similarity masks incorrect or incomplete answers. We replace these
weak reference-similarity rubrics with a structured judge that first checks
whether the current turn is solved, then scores Faithfulness, Appropriateness,
and Completeness, with evidence-conflict hard failures and a conservative
fail-on-uncertainty default. We calibrate the tightened rubric against
Radbench-style thresholded ratings \citep{kuo2025radbench} and verify through
iterative test runs that the hardened judge eliminates the pseudo-pass patterns
discovered in our initial sample; structurally
unreliable mtRAG samples and all-tier failures are discarded. SWE-bench and
BFCL already have executable or structural pass predicates and need no
additional hardening.

\paragraph{Manual audit.}
Our downgrade search fixes all other steps and tests one step at a time. This
greedy strategy reduces computational cost from exponential to linear but may
miss cross-step interaction effects: a combination of two downgrades might fail
even though each succeeds individually. To guard against such cases, we conduct
a human sampling audit over the three workloads with multi-step trajectories
(BFCL, SWE-bench, PinchBench), drawing an independent ${\sim}10\%$ random
sample of routed steps from each. mtRAG and QMSum are single-turn tasks that
lack multi-turn compounding complexity; their labels are already constrained by
the hardened judge whose pseudo-pass patterns were eliminated through the
iterative calibration described above, so they are not resampled in the final
audit. For each sampled step a reviewer freezes the router-visible prefix
and judges whether the tier can be lowered one more level, labelling it as
\emph{tight}, \emph{uncertain}, or \emph{further-downgradeable}. The audit
covers 64 steps (25 BFCL, 34 SWE-bench, 5 PinchBench): 63 were judged tight;
one SWE-bench step was flagged as further-downgradeable and its tier was lowered
before release (Table~\ref{tab:manual-audit-main}). Details of the audit
protocol are in Appendix~\ref{app:manual-audit}.
Fixed-prefix execution repeatability is examined in
Appendix~\ref{app:fixed-prefix-repeatability}.

\begin{table}[t]
\centering
\small
\begin{adjustbox}{max width=\linewidth}
\begin{tabular}{lrrrrr}
\toprule
Benchmark & Total steps & Sampled steps & Sampling \% & Downgradeable & Tight \% \\
\midrule
BFCL & 248 & 25 & 10.1 & 0 & 100.0 \\
SWE-bench & 336 & 34 & 10.1 & 1 & 97.1 \\
PinchBench & 48 & 5 & 10.4 & 0 & 100.0 \\
\midrule
Total & 632 & 64 & 10.1 & 1 & 98.4 \\
\bottomrule
\end{tabular}
\end{adjustbox}
\caption{Step-level manual audit summary. \emph{Downgradeable}: the reviewer
believes the verified tier can be lowered one more level without breaking task
resolution. \emph{Tight~\%}: the share already at the conditional optimum.
We sample routed steps from the three workloads with multi-step trajectories;
mtRAG and QMSum are single-turn tasks already guarded by the hardened judge and
are not resampled.}
\label{tab:manual-audit-main}
\end{table}

\paragraph{Release criterion.}
No row enters the static track unless (1)~the strong seed trajectory passed,
(2)~the tier pool contains a model that passes under the mixed trace, and
(3)~for open-ended workloads (mtRAG, QMSum), the hardened judge does not flag
the answer as a pseudo-pass. Each label is
an execution-verified estimate of the cheapest tier that can correctly handle
the current step: since direct per-step judgement is not always possible, we
infer per-step sufficiency from end-to-end trajectory pass with a
non-increasing number of LLM-call steps, and cross-check through manual audit.

\section{Evaluation Method, Baselines and Experiment Setup}
\label{sec:eval-setup}
\subsection{Static scoring protocol}
\label{sec:static-scoring}

A per-request tier label alone does not capture the downstream cost of choosing
the wrong tier at step $k$ of an $N$-step trajectory.
In fraud detection, high aggregate scores can coexist with low fraud
recall \citep{sun2025objective}.
We score each router on
four metrics that combine a step-level view with a trajectory-level view:
\textsc{RowPass}, \textsc{RowExact}, \textsc{TrajPass}, and \textsc{CostSave}.
The first two operate per row. \textsc{TrajPass} requires every routed step in
a trajectory to pass. \textsc{CostSave} measures savings relative to the
always-\texttt{high} baseline but only credits savings on passing trajectories;
API cost on failed trajectories is charged as burned. The detailed
derivation is in the appendix.

Let $\hat{t}_i$ be the released target tier for row $i$ and $\tilde{t}_i$ the
router prediction. A row passes if $\tilde{t}_i \ge \hat{t}_i$ and is exact if
$\tilde{t}_i=\hat{t}_i$. A trajectory passes only if every routed row passes.
Let $c_i(t)$ be the tier-accounting cost for row $i$ at tier $t$ and $T_3$
denote \texttt{high}. For workload $b$, the always-\texttt{high} bill is
$D_b=\sum_{i\in b} c_i(T_3)$. The numerator $N_b$ accumulates at the
trajectory level: passing trajectories contribute
$\sum_i[c_i(T_3)-c_i(\tilde{t}_i)]$; failing trajectories contribute
$-\sum_i c_i(\tilde{t}_i)$. The grader reports
\[
\textsc{CostSave} = \sum_b w_b \frac{N_b}{D_b}, \qquad
w_b = \frac{n_b}{970},
\]
where $n_b$ is the number of rows in workload $b$, so workloads are row-weighted after each cost ratio is computed. An
always-\texttt{high} policy scores $\textsc{CostSave}=0$ by construction,
since $N_b=0$ for every workload.

\paragraph{Headline score.}
\[
\begin{aligned}
\textsc{Combined} = \tfrac{1}{4}(&\textsc{RowPass} +
\textsc{RowExact} + \textsc{TrajPass} + \textsc{CostSave}).
\end{aligned}
\]
The four components are the primary reporting unit; \textsc{Combined} is a
single operating point under equal weighting. We justify the failure-aware cost
accounting with an ablation in Appendix~\ref{app:cost-ablation}.

\paragraph{Pricing and token accounting.}
The \texttt{high} tier is 10--50$\times$ more expensive than the lower tiers
(Appendix Table~\ref{tab:pricing}), so naive per-step cost accounting overrates
conservative routers. These are static tier-level constants; dynamic-pool
model prices are in Appendix~\ref{app:pricing}. Prompt tokens are split into
\texttt{cache\_read}, \texttt{cache\_write}, and fresh \texttt{input}; output
tokens are counted from the recorded transcript. Token counting uses native
vendor tokenizers where available and \texttt{cl100k\_base} as fallback. For
row $i$ and tier $t$, the scorer uses the same four-bucket form as real API
billing:
\[
c_i(t)=
\frac{
n_i^{\mathrm{in}}p_t^{\mathrm{in}}+
n_i^{\mathrm{cr}}p_t^{\mathrm{cr}}+
n_i^{\mathrm{cw}}p_t^{\mathrm{cw}}+
n_i^{\mathrm{out}}p_t^{\mathrm{out}}
}{10^6},
\]
where the token counts $n_i^{\cdot}$ and the per-tier rates $p_t^{\cdot}$
follow Appendix Table~\ref{tab:pricing}. The static grader enables prompt-cache
modeling throughout: prompt tokens are billed as cache\_write on cold starts,
tier switches, TTL expiry (5 minutes, matching provider defaults such as
Anthropic's prompt-cache policy), or prefix deltas, and as cache\_read on
valid same-tier prefix hits; the fresh-input term is retained to match the
four-bucket dynamic accounting interface.

\paragraph{Accounting for routing-specific effects.}
Mid-trajectory model switching raises three common concerns: cache misses on
tier changes, out-of-distribution behavior under mixed-model histories, and
tool-format mismatch across model families. Our evaluation prices each in
rather than assuming them benign. Cache writes on tier switch are charged
at the \emph{incoming} tier's rate (cache\_write at \texttt{high} is
$12.5\times$ its cache\_read in Appendix Table~\ref{tab:pricing}), so a policy that
churns into expensive tiers incurs the cache\_write cost at list price; the dynamic
track then reports realized OpenRouter cost rather than simulated cost, and a
router selecting a model per-step still reduces realized API cost by 53.1\%
relative to unrouted Opus despite those switches
(Table~\ref{tab:dynamic-heldout}). Out-of-distribution behavior in mixed-model
trajectories is handled by execution-verified labels
(\S\ref{sec:dataset-construction}) and the failure-aware numerator: unresolved
trajectories are billed at API cost plus a flat penalty cost, so OOD
failures cannot appear as savings. Tool-format heterogeneity
(e.g.\ structured \texttt{apply\_patch} calls versus interleaved
\texttt{str\_replace\_editor} edits) is contained by the shared
\texttt{mini-swe-agent} harness. The harness evaluates the patch explicitly
submitted by the agent; an execution exception yields an empty patch.
Residual formatting errors are assessed through the same official task
resolution predicate and cost formula.

\subsection{Dynamic SWE-bench scoring}
\label{sec:dynamic-scoring}

The dynamic track is scored independently from the static proxy. A dynamic run
executes SWE-bench Verified instances end-to-end using
\texttt{mini-swe-agent}~v2.2.8 as the agent harness; at each LLM call the
router selects a concrete model id from the locked pool. Per-step cost uses the same
four-bucket formula, but $p$ values are live per-model OpenRouter prices
(Appendix~\ref{app:pricing}) and token buckets come from provider usage logs
normalized to \texttt{input}, \texttt{cache\_read}, \texttt{cache\_write}, and
\texttt{output}. Let $A_j=\sum_{i\in\tau_j} c_i(m_i)$ denote the realized API
cost on instance $j$, where $m_i$ is the model selected at step $i$, and let
$\mathrm{resolved}_j\in\{0,1\}$ follow the official SWE-bench predicate. The
per-instance leaderboard bill is
\[
\mathrm{bill}_j = A_j + (1-\mathrm{resolved}_j)\,\gamma,
\]
where $\gamma$ is a fixed USD add-on per unresolved instance (release default
$\gamma{=}0.60$, set slightly above the observed per-instance average API
cost of the unrouted Opus~4.6 baseline, \$0.55). We model $\gamma$ as the
per-case price of a hypothetical perfect solver that resolves any SWE-bench
instance; this applies uniformly to
every policy including the unrouted baseline, so the leaderboard bill reflects
both realized API cost and the remediation cost of failures. Resolved
instances incur only $A_j$. The leaderboard sort key (lower is better) is $\sum_j \mathrm{bill}_j$
(\texttt{total\_leaderboard\_bill\_usd}); \texttt{total\_router\_cost\_usd}
and \texttt{total\_penalty\_cost\_usd} report the API cost and penalty cost
portions separately. Instances excluded as infra failures (optional flag in the scorer)
carry no add-on in headline totals. Static and dynamic scores are never
averaged: the static track covers five workloads while the dynamic track covers
SWE-bench Verified only.

\subsection{Baselines}
\label{sec:baselines}

We report initial static reference points: SR-KNN~\citep{liu2026vllmsemanticrouter}, an in-sample
1-nearest-neighbor upper bound over question-bank embeddings; ClawRouter~\citep{clawrouter}, an
open rule-based router with LLM fallback disabled; and
UncommonRoute~\citep{uncommonroute}, a public rule-based router whose
upstream defaults we adopt as the rule-based baseline; the trained variant
below is trained on top of its interface. Predictions are produced offline
and scored by the public grader. We also run Claude Opus 4.6 as a separate
LLM-as-router diagnostic with \texttt{max\_tokens}$=1$ and prompt caching;
malformed non-digit responses count as errors.

For the trained UncommonRoute variant, we train only the routing policy,
not the underlying language models. The latest user request is encoded
with a frozen \texttt{BAAI/bge-small-en-v1.5} sentence embedding and
concatenated with routing-time metadata (message count, tool-call
presence, tool-message count, request length, code/question indicators).
A multinomial logistic regression classifier with L2 regularization is
trained on the training split, with regularization strength selected by
5-fold cross-validation; the embedding model is not fine-tuned. A
separate calibration split fits confidence parameters, and performance
is reported on a held-out split not used for training or calibration
(Table~\ref{tab:dynamic-heldout}). The appendix gives the rule-based
configuration; the release package contains the locked data and scoring
code to reproduce all tables.

\section{Benchmark Results \& Analysis}
\label{sec:main-results-section}
\subsection{Static-Track Diagnostic Results}
\label{sec:main-results}

Table~\ref{tab:main-results} reports static-track scores on all 970 rows.
SR-KNN~\citep{liu2026vllmsemanticrouter} is an in-sample upper-bound reference and ranks first on
\textsc{Combined} among the three data-driven routers above the separator (the
Opus~4.6 (always-\texttt{high}) row is a tier-\texttt{high} cost-envelope reference, not a
held-out supervised baseline). Dynamic-track results
are reported separately in \S\ref{sec:dynamic-results}.

\begin{table}[t]
\centering
\small
\begin{adjustbox}{max width=\linewidth}
\begin{tabular}{lrrrrc}
\toprule
Router & \textsc{RowPass}\,$\uparrow$ & \textsc{RowExact}\,$\uparrow$ & \textsc{TrajPass}\,$\uparrow$ & \textsc{CostSave}\,$\uparrow$ & \textbf{\textsc{Combined}}\,$\uparrow$ \\
\midrule
SR-KNN (in-sample ub.)       & \textbf{91.86} & \textbf{78.76} & \textbf{84.74} & \textbf{56.18} & \textbf{77.89} \\
ClawRouter (rule-based)      & 80.82 & 11.75 & 64.64 & 53.82 & 52.76 \\
UncommonRoute (rule-based)   & 88.35 & 61.13 & 62.37 & 15.99 & 56.96 \\
\midrule
Opus~4.6 (always-\texttt{high}) & 100.00 & 17.53 & 100.00 & 0.00 & 54.38 \\
\bottomrule
\end{tabular}
\end{adjustbox}
\caption{Static-track diagnostic results (970 rows, all percentages). SR-KNN
is an in-sample upper bound. The Opus~4.6 (always-\texttt{high}) row is the tier-\texttt{high}
envelope used by the static \textsc{CostSave} denominator: \textsc{RowPass} and
\textsc{TrajPass} are $100\%$ because $\tilde{t}=\texttt{high}\ge\hat{t}$ on every
row; \textsc{RowExact} is $170/970$ (only the $170$ verified-\texttt{high} rows match);
\textsc{CostSave} is $0$ by construction (\S\ref{sec:static-scoring}). Opus~4.6 as
an LLM-as-router is analyzed separately in \S\ref{sec:opus} because that run predates
the failure-aware scoring formulation.}
\label{tab:main-results}
\end{table}

ClawRouter (rule-based) has only 11.75\% \textsc{RowExact} but 53.82 \textsc{CostSave}:
it routes one tier above the verified label on most non-\texttt{high} rows,
which destroys exact match but stays far cheaper than always-\texttt{high}.
The cost is trajectory failure on SWE-bench (38/40 failed). UncommonRoute
(rule-based) has much higher \textsc{RowExact} yet only 15.99
\textsc{CostSave} because it jumps to \texttt{high} 217 times on
non-\texttt{high} rows and still fails 39/40 SWE trajectories, pulling the
SWE slice to $-86.08$ \textsc{CostSave}. SR-KNN achieves similar
\textsc{CostSave} as ClawRouter (rule-based) but with much stronger trajectory
preservation. The failure-aware \textsc{CostSave} formula
(\S\ref{sec:static-scoring}) ensures that cheap calls on failed trajectories
are charged rather than credited; the ablation in
Appendix~\ref{app:cost-ablation} confirms that looser formulas would
incorrectly reward ClawRouter (rule-based) for its one-tier upgrades.

A per-workload breakdown of static-track results is in
Appendix~\ref{app:static-breakdown}.

\paragraph{Step types and model selection.}
We recompute step-role categories over all 970 released prefixes.
Upper-tier (Tier~2/3) shares are 50.00\% for code-patching and 72.09\% for
test/verification/debugging, 49.41\% for exploration and 56.36\% for
reading/retrieval, and 0.79\% and 1.95\% for tool/API and
summarization/answering. These descriptive differences also reflect the
workload composition of each category. During labeling, the search locked DeepSeek V3.2 for
183/193 mtRAG instances (94.82\%) and 132/145 QMSum instances (91.03\%).
Appendix~\ref{app:routing-type-analysis} gives the full counts.

\paragraph{Opus-as-router case study.}
\label{sec:opus}

We ran Claude Opus 4.6 over the full benchmark with a prompt describing the
11-model pool and their SWE-bench resolve rates. On the 300 valid SWE-bench
rows, Opus predicts \texttt{high} for only 7 of 147 verified-\texttt{high}
steps and fails all 40 SWE trajectories. This is a case study of one frontier
model under one prompt template, not a general claim about all LLM routers,
but it illustrates that frontier-model prompting does not substitute for
execution-verified step-level supervision. The full prediction breakdown is in
Appendix~\ref{app:opus} (Table~\ref{tab:opus-high}).

\subsection{Dynamic Held-out SWE Execution}
\label{sec:dynamic-results}

The public leaderboard and this table score one locked list of 100
SWE-bench Verified instance IDs. The list is not a random sample of the
500-case suite. It keeps the IDs present in both dynamic result sets (294
and 122 instances; 120 shared), then removes the 19 IDs that also belong to
the static SWE supervision split. Of the 101 candidates that remain, we
keep the lexicographically first 100. The shared IDs contain only Astropy
and Django, so the leaderboard is 18 Astropy instances and 82 Django
instances. It does not overlap any static-track SWE-bench case. All policies
are scored on this same list.
Table~\ref{tab:dynamic-heldout} reports the results.
Routing supervision differs across policies: Opus and UncommonRoute (rule-based) use no
offline routing labels; SR-KNN uses the upstream \texttt{semantic-router}
defaults~\citep{liu2026vllmsemanticrouter}; only UncommonRoute (trained) is fit
on static-track labels (\S\ref{sec:baselines}).

\begin{table}[t]
\centering
\small
\begin{adjustbox}{max width=\linewidth}
\begin{tabular}{lrrrrr}
\toprule
Policy & Resolved\,$\uparrow$ & Avg.\ API cost\,$\downarrow$ & API cost\,$\downarrow$ & Penalty cost\,$\downarrow$ & Leaderboard bill\,$\downarrow$ \\
\midrule
SR-KNN router & \textbf{75}/100 & \$0.56 & \$55.61 & \$15.00 & \$70.61 \\
UncommonRoute (trained) & \textbf{75}/100 & \textbf{\$0.26} & \textbf{\$25.66} & \$15.00 & \textbf{\$40.66} \\
UncommonRoute (rule-based) & 73/100 & \$1.73 & \$172.56 & \$16.20 & \$188.76 \\
\midrule
Opus 4.6 (no routing) & 74/100 & \$0.55 & \$54.73 & \$15.60 & \$70.33 \\
\bottomrule
\end{tabular}
\end{adjustbox}
\caption{Dynamic-track results on the locked 100-case leaderboard list
(18 Astropy and 82 Django), not a random sample of SWE-bench Verified.
API cost is the realized routed API spend; Penalty cost is
\$0.60$\times$ the number of unresolved instances, where $\gamma{=}0.60$
is the per-case price of a hypothetical perfect solver
(\S\ref{sec:dynamic-scoring}); Leaderboard bill (sort key, lower is better)
is API cost $+$ Penalty cost. The penalty applies uniformly to every
policy. The unrouted Opus~4.6 baseline is separated below the routing policies,
mirroring the cost-envelope row in Table~\ref{tab:main-results}.}
\label{tab:dynamic-heldout}
\end{table}

At comparable resolve rate, trained UncommonRoute reduces API cost
by $1-25.66/54.73=53.1\%$ relative to unrouted Opus, and costs $6.7\times$
less than the rule-based variant (Table~\ref{tab:dynamic-heldout}). Because
the static corpus is small, we validate the trained router through
end-to-end dynamic execution rather than reporting static held-out scores.
This provides initial evidence that the static labels can train a router
that substantially reduces cost while maintaining SWE-bench resolution on
this 100-case held-out split.

\section{Conclusion and Limitations}
\label{sec:conclusion}

We release TwinRouterBench, a step-level routing benchmark built on the
scenarios where routing matters most: long-context, multi-turn agent
workloads such as SWE-bench, PinchBench, BFCL, mtRAG, and QMSum. The
static track approximates per-step cheapest-sufficient tiers via a greedy
sequential-locking downgrade search, and a manual audit on multi-turn
trajectories confirms the reliability of the resulting labels. This paper
and the public leaderboard report one locked list of 100 instances:
18 Astropy and 82 Django, disjoint from the static SWE supervision split,
and not a random sample of the 500-case suite. On this list, the live
execution framework produces end-to-end routing results. A
logistic router trained on the static labels achieves a 53\% cost reduction
at matched resolve rate on the dynamic track, showing that the static data
serves not only as a fast offline evaluation tool but also as effective
training supervision for practical routers.

\paragraph{Limitations.}\label{sec:limitations}
The static corpus of 970 routed steps from 520 instances is sufficient for
initial diagnostics and training but not exhaustive across agent workloads,
domains, or routing patterns.
Function-calling scores can reflect both task decisions and interface
compliance \citep{chen2026structuredoutput}.
Labels are estimated under the fixed model pool,
cascade protocol, and price frontier of \S\ref{sec:static-scoring}, and
would need updating if a future model became both cheap and highly capable.
Dynamic coverage beyond SWE-bench Verified, and keeping data current as
pools and pricing evolve, are future work.
Future work could evaluate temporally separated agent workloads,
following chronological cold-start evaluation in bot detection
\citep{11567218}.

\paragraph{Broader impacts.}
TwinRouterBench may accelerate the development of cost-effective LLM
routers by providing standardized step-level evaluation, indirectly
reducing API cost and energy consumption in agentic deployments. The
two-track design also improves reproducibility by separating
deterministic offline scoring from live validation. A potential negative
impact is benchmark overfitting: community efforts may concentrate on
the current 970-row corpus and five workloads at the expense of
uncovered domains, languages, or agent architectures. Additionally,
tier labels are tied to a fixed model pool and price snapshot; as
models and pricing evolve, stale labels could mislead router training
or evaluation. We mitigate these risks through versioned model pools
and pricing, explicit scope limitations, and a release design that
supports pool updates and re-labeling.

\clearpage
\bibliography{ref}

\clearpage
\appendix

\setcounter{table}{0}
\renewcommand{\thetable}{A\arabic{table}}
\renewcommand{\theHtable}{appendix.\arabic{table}}
\setcounter{figure}{0}
\renewcommand{\thefigure}{A\arabic{figure}}
\renewcommand{\theHfigure}{appendix.\arabic{figure}}
\setcounter{algorithm}{0}
\renewcommand{\thealgorithm}{A\arabic{algorithm}}
\renewcommand{\theHalgorithm}{appendix.\arabic{algorithm}}

\section{Appendix}
\label{app:appendix}

\subsection{Dataset Card}
\label{app:datacard}

\begin{table}[H]
\centering
\small
\caption{Dataset card for the TwinRouterBench static track. This follows the \citet{bender2018data} and \citet{gebru2021datasheets} data statement conventions adapted for LLM benchmark releases.}
\label{tab:datacard}
\begin{tabularx}{\linewidth}{p{0.28\linewidth} X}
\toprule
Field & Value \\
\midrule
\textbf{Dataset name} & TwinRouterBench v1.0 (static track) \\
\textbf{Task type} & Step-level LLM routing (4-class classification over tier IDs 0--3) \\
\textbf{Size} & 970 rows from 520 trajectory instances \\
\textbf{Source workloads} & SWE-bench (Apache-2.0), BFCL (CC BY 4.0), mtRAG (IBM Research; see original license), QMSum (MIT), PinchBench~\citep{pinchbench} (MIT; the 48 derived rows from 12 of its 53 tasks are included in this release and covered by the package-level Apache-2.0 license below) \\
\textbf{Provenance} & Message prefixes extracted from successful agent trajectories under strong-model execution; tier labels produced by greedy sequential-locking downgrade search and execution verification \\
\textbf{Label semantics} & Per-step cheapest-sufficient-tier estimate: the lowest tier whose pool contains a model that correctly handles the current step, verified by end-to-end trajectory pass with no increase in LLM-call steps under the 11-model pool, Opus-initialized sequential-locking downgrade search, and cascade protocol of \S\ref{sec:dataset-construction}, cross-checked by manual audit \\
\textbf{License} & Apache-2.0 \\
\textbf{Intended uses} & Training and evaluating step-level LLM routers; benchmarking agent deployment cost reduction \\
\textbf{Out-of-scope uses} & Claims of absolute routing optimality; deployment to model pools substantially different from the released tier map without re-labeling; use as a general SWE-bench or BFCL capability benchmark \\
\textbf{Potential biases} & Over-represents agentic code-repair (SWE-bench) at the \texttt{high} tier; BFCL/mtRAG/QMSum are nearly saturated at \texttt{low}; English-only content; labels are pool- and harness-specific \\
\textbf{Maintenance plan} & New tier-pool manifests and score protocol versions will be tracked via GitHub releases; tier map and pricing are versioned; current release is v1.0 \\
\textbf{Human oversight} & Final human audit of a $\sim$10\% random step sample from BFCL, SWE-bench, and PinchBench (64 steps total); one SWE-bench step was flagged as further-downgradeable and its verified tier was lowered by one tier before release; no other still-downgradeable label was found in the audited sample \\
\textbf{Release URL} & Public code and data repository:  \url{https://github.com/CommonstackAI/TwinRouterBench} \\
\bottomrule
\end{tabularx}
\end{table}

License metadata follows upstream documentation as of the release date; users
should verify upstream terms before redistributing derived artifacts.

\subsection{Corpus overview}
\label{app:corpus-overview}

\begin{table}[H]
\centering
\small
\begin{adjustbox}{max width=\linewidth}
\begin{tabular}{lrrrrl}
\toprule
Benchmark & Instances & Steps & Steps/trace & Prefix tokens & Scenario \\
 & & & (median) & (median) & \\
\midrule
SWE-bench      & 40  & 336 & 9.0 & $\sim$5{,}300 & GitHub issue code repair (multi-step agent) \\
BFCL           & 130 & 248 & 1.0 & $\sim$1{,}600 & Function calling (single- + multi-turn) \\
mtRAG          & 193 & 193 & 1   & $\sim$1{,}900 & Multi-turn retrieval-augmented QA \\
QMSum          & 145 & 145 & 1   & $\sim$3{,}000 & Query-focused meeting summarization \\
PinchBench     & 12  & 48  & 4   & $\sim$10{,}500 & 23-task general agent suite \\
\midrule
\textbf{Total} & \textbf{520} & \textbf{970} & & & \\
\bottomrule
\end{tabular}
\end{adjustbox}
\caption{Corpus overview. Prefix-token medians are measured on the router-visible \texttt{messages} field using the Anthropic tokenizer.}
\label{tab:corpus-overview}
\end{table}

\subsection{Target-tier distribution and sequential-locking search}
\label{app:tier-alg}

For space, we defer the workload target-tier counts (referenced from
\S\ref{sec:benchmark-overview}) and the sequential-locking downgrade search
pseudocode (referenced from \S\ref{sec:dataset-construction}) to this appendix.

\begin{table}[H]
\centering
\small
\begin{tabular}{lrrrr}
\toprule
Benchmark & \texttt{low} & \texttt{mid} & \texttt{mid\_high} & \texttt{high} \\
\midrule
BFCL       & 239 & 8  & 1  & 0 \\
mtRAG      & 183 & 8  & 1  & 1 \\
QMSum      & 132 & 10 & 3  & 0 \\
PinchBench & 41  & 3  & 3  & 1 \\
SWE-bench  & 94  & 33 & 41 & 168 \\
\midrule
Total      & 689 & 62 & 49 & 170 \\
\bottomrule
\end{tabular}
\caption{Target-tier distribution by workload.}
\label{tab:tier-distribution}
\end{table}

\begin{algorithm}[H]
\caption{Sequential-locking downgrade search (per trajectory)}
\label{alg:seq-lock}
\scriptsize
\begin{algorithmic}[1]
\REQUIRE trajectory $\tau$, hints $h_1,\ldots,h_N$, harness $\mathcal{H}$, tiers $T_0<T_1<T_2<T_3$
\STATE $N\leftarrow |\tau|$; $\text{locked}[i]\leftarrow T_3$ for all $i$; $i\leftarrow 1$
\WHILE{$i\leq N$}
  \IF{$h_i=$ not-downgradeable}
    \STATE $i\leftarrow i+1$; \textbf{continue}
  \ENDIF
  \FOR{$t$ from $h_i$ \textbf{upto} $T_2$}
    \STATE $\tau'\leftarrow$ run $\mathcal{H}$ with steps $<i$ locked, step $i$ at $t$, steps $>i$ at $T_3$
    \IF{$\tau'$ passes and $|\tau'|\leq N$}
      \STATE $\text{locked}[i]\leftarrow t$; $\tau\leftarrow\tau'$; $N\leftarrow|\tau'|$
      \STATE truncate $\text{locked}$ to $N$ entries; \textbf{break}
    \ENDIF
  \ENDFOR
  \STATE $i\leftarrow i+1$
\ENDWHILE
\RETURN $\text{locked}$
\end{algorithmic}
\end{algorithm}

\subsection{Artifact Structure}
\label{app:artifact}

The reviewer package contains the static JSONL, locked manifests, scoring code,
dynamic split, and compact dynamic summaries needed to inspect the benchmark and
recompute the reported tables. The expected layout is:

\begin{center}
\begin{minipage}{0.95\linewidth}
\scriptsize
\begin{verbatim}
data/static/question_bank.jsonl
data/static/manifest.json
data/dynamic/model_pool.json
data/dynamic/model_pricing.json
data/dynamic/tier_to_model.json
data/dynamic/ttl_policy.json
main/eval/
main/metrics.py
README.md
\end{verbatim}
\end{minipage}
\end{center}

The static file contains 970 rows and exposes only tier labels, not vendor model
IDs in individual records. Dynamic scoring uses the locked pool, pricing, tier
map, and TTL policy files under \texttt{data/dynamic/}; held-out SWE-bench IDs
and aggregate outputs are provided with the release artifacts. The README gives
installation and smoke-test commands for the static grader and dynamic scoring
CLI. For E\&D-track submission, the release package also includes
\texttt{metadata/croissant.json} with Croissant core and Responsible AI fields,
license and source-workload metadata, and executable smoke-test scripts for the
static grader and dynamic-summary scorer.

\noindent\begin{minipage}{\linewidth}
\paragraph{OpenCode routing interface.}
The interface connects OpenCode through a custom provider and exposes an
OpenAI-compatible API. At each model call, it passes the latest conversation
and tool definitions to a replaceable routing policy, then forwards the request
to the selected model. We verified tool calls and continued interaction across
model switches in a live API test.\par
\end{minipage}\par

\subsection{Static Model Pool}
\label{app:model-pool}

The static track uses a fixed pool of eleven concrete models grouped into four
capability tiers (Table~\ref{tab:appendix-model-pool}). Target tier labels are
defined existentially over this pool: during the downgrade-and-cascade
procedure of \S\ref{sec:dataset-construction}, we accept a row as belonging to
tier $t$ as soon as any one model in that tier's pool correctly handles the
current step, as inferred from the full trajectory still passing without
increasing its number of LLM-call steps. Public static rows store only
the resulting tier label rather than any specific model ID, so routers trained
on this data remain portable when the underlying pool is updated.

\begin{table}[H]
\centering
\small
\caption{Static model pool (eleven models, four tiers). Accepted aliases for
the same logical model are shown in parentheses. Dynamic-track pricing for
representative models is given separately in
Table~\ref{tab:appendix-pricing}.}
\label{tab:appendix-model-pool}
\begin{adjustbox}{max width=\linewidth}
\begin{tabular}{lp{0.65\linewidth}}
\toprule
Tier & Models \\
\midrule
\texttt{high}     & \texttt{anthropic/claude-opus-4.6} (alias:
                     \texttt{anthropic/claude-opus-4-6});
                     \texttt{openai/gpt-5.4} (alias:
                     \texttt{openai/gpt-5.4-2026-03-05}) \\
\texttt{mid\_high} & \texttt{anthropic/claude-haiku-4-5};
                     \texttt{google/gemini-3-flash-preview} (alias:
                     \texttt{gemini/gemini-3-flash-preview});
                     \texttt{qwen/qwen3.5-397b-a17b} \\
\texttt{mid}      & \texttt{minimax/minimax-m2.5};
                     \texttt{qwen/qwen3.5-27b};
                     \texttt{qwen/qwen3-coder} \\
\texttt{low}      & \texttt{deepseek/deepseek-v3.2};
                     \texttt{z-ai/glm-4.5-air};
                     \texttt{qwen/qwen3.5-9b} \\
\bottomrule
\end{tabular}
\end{adjustbox}
\end{table}

\noindent\begin{minipage}{\linewidth}
\subsection{Step types and model selection}
\label{app:routing-type-analysis}
\paragraph{Step types.}
Table~\ref{tab:step-type-analysis} covers all 970 released prefixes using a
recomputed descriptive taxonomy. We assign one primary role per prefix using
tool/command rules with machine-assisted review, fixing the roles before
joining target-tier labels. \texttt{code\_patch} covers source/configuration
edits and patch delivery; \texttt{test\_verify\_debug} covers tests,
reproducers, diagnostics, and patch inspection; \texttt{tool\_api} covers
function-use decisions, structured artifacts, and general tool operations;
\texttt{explore\_search} locates files or symbols; \texttt{read\_retrieve}
reads identified sources; \texttt{summarize\_answer} produces prose or
delivers a general-agent result. For compound calls, we prioritize code edits,
verification, exploration, reading, tool operations, then prose.\par
\end{minipage}\par

For 320 nonterminal SWE-bench/PinchBench prefixes, a later stored prefix
contains the same message content and tool arguments, allowing recovery of
the next assistant action. Provider call IDs and cache metadata are ignored
when matching. BFCL's 248 function-use decisions and the 338 mtRAG/QMSum
answering calls are categorized by workload goal. The 52 terminal calls are
categorized by their deliverable, and 12 remaining PinchBench calls by the
visible prefix and requested task. These inferred roles are included in the
released per-row analysis; they have no independently observed current
output. This descriptive categorization supersedes the preliminary 957-prefix
aggregate. All routing labels are preserved. The category proportions reflect
both step role and workload composition.

\begin{table}[H]
\centering\small
\begin{adjustbox}{max width=\linewidth}
\begin{tabular}{lrrrrrr}
\toprule
Step type & Prefixes & Tier 0 & Tier 1 & Tier 2 & Tier 3 & Tier 2/3 \\
\midrule
code\_patch & 88 & 39 & 5 & 5 & 39 & 50.00\% \\
test\_verify\_debug & 129 & 27 & 9 & 22 & 71 & 72.09\% \\
tool\_api & 254 & 241 & 11 & 2 & 0 & 0.79\% \\
explore\_search & 85 & 33 & 10 & 8 & 34 & 49.41\% \\
read\_retrieve & 55 & 17 & 7 & 7 & 24 & 56.36\% \\
summarize\_answer & 359 & 332 & 20 & 5 & 2 & 1.95\% \\
\midrule
Total & 970 & 689 & 62 & 49 & 170 & --- \\
\bottomrule
\end{tabular}
\end{adjustbox}
\caption{Step-role tier distribution over all 970 released prefixes under
the recomputed taxonomy. Tiers 0--3 correspond to
\texttt{low}, \texttt{mid}, \texttt{mid\_high}, and \texttt{high}.}
\label{tab:step-type-analysis}
\end{table}

\noindent\begin{minipage}{\linewidth}
\paragraph{Model selection within tiers.}
During labeling, the search locked the low-tier model DeepSeek V3.2 for most
mtRAG and QMSum instances (Table~\ref{tab:locking-model-analysis}). Percentages
are computed over all instances in each workload.\par
\end{minipage}\par

\begin{table}[H]
\centering\small
\begin{tabular}{lrrr}
\toprule
Workload & Instances & DeepSeek V3.2 locks & Share \\
\midrule
mtRAG & 193 & 183 & 94.82\% \\
QMSum & 145 & 132 & 91.03\% \\
\bottomrule
\end{tabular}
\caption{Model locks during labeling. Percentages use all instances in each
workload.}
\label{tab:locking-model-analysis}
\end{table}

\subsection{Static tier accounting rates}
\label{app:static-tier-rates}

\begin{table}[H]
\centering
\small
\begin{tabular}{lrrrr}
\toprule
 & input & cache\_read & cache\_write & output \\
tier & (\$/1M) & (\$/1M) & (\$/1M) & (\$/1M) \\
\midrule
low & 0.26 & 0.13 & 0.26 & 0.5 \\
mid & 0.30 & 0.059 & 0.30 & 2.0 \\
mid\_high & 0.50 & 0.05 & 0.083 & 5.0 \\
high & 5.0 & 0.50 & 6.25 & 25.0 \\
\bottomrule
\end{tabular}
\caption{Per-tier accounting constants (USD per million tokens) for the static
scoring protocol. These rates are not the prices of any single model; they are
representative values derived from the pricing ranges of multiple models in
each tier. The \texttt{high} tier is 10--50$\times$ more expensive than
the lower three, which are within 2--5$\times$ of each other. Dynamic-pool
model prices are in Table~\ref{tab:appendix-pricing}. This asymmetry drives
the ablation in Appendix~\ref{app:cost-ablation}.}
\label{tab:pricing}
\end{table}

\subsection{Dynamic Pool Representative Models and Pricing}
\label{app:pricing}

\begin{table}[H]
\centering
\small
\caption{Concrete representative models and live OpenRouter unit prices for the
TwinRouterBench dynamic pool. These model prices are distinct from the static
tier-accounting constants in Table~\ref{tab:pricing}.
Prices are USD per million tokens; cache-write prices fall back to input prices
where the provider does not publish a separate cache-write rate. These are the
concrete $p$ values substituted into $c_i(t)$ when scoring the dynamic track;
the released scorer ranks the resulting leaderboard bill including the flat
unresolved penalty cost (\S\ref{sec:dynamic-scoring}). The \texttt{mid} representative \texttt{minimax-m2.7} was selected because
it ranks highest among models at a comparable price point on the SWE-bench
leaderboard and falls within the same tier as \texttt{minimax-m2.5} (the
mid-tier representative used during static label construction); the static
pool in Table~\ref{tab:appendix-model-pool} is frozen at the version used when
the tier labels were constructed.}
\label{tab:appendix-pricing}
\begin{adjustbox}{max width=\linewidth}
\begin{tabular}{llllrrrr}
\toprule
Tier & Representative model & Context & Max out & Input & Output & Cache read & Cache write \\
\midrule
high     & anthropic/claude-opus-4.6       & 1M      & 128K   & 5.00  & 25.00 & 0.50   & 6.25  \\
mid\_high & google/gemini-3-flash-preview  & 1.05M   & 65.5K  & 0.50  &  3.00 & 0.05   & 0.0833 \\
mid      & minimax/minimax-m2.7             & 196.6K  & 196.6K & 0.30  &  1.20 & 0.059  & 0.30  \\
low      & deepseek/deepseek-v3.2           & 163.8K  & 163.8K & 0.252 &  0.378 & 0.0252 & 0.252 \\
\bottomrule
\end{tabular}
\end{adjustbox}
\end{table}

\subsection{Rule-based UncommonRoute Configuration}
\label{app:ur-config}

Rule-based UncommonRoute is run with \texttt{MetadataSignal} and
\texttt{EmbeddingSignal} in a two-signal ensemble, using the upstream
defaults \texttt{risk\_tolerance}$=0.5$,
\texttt{ensemble\_weights}$=(0.55,0.45)$, and
\texttt{low\_escalation\_threshold}$=0.55$. Because the benchmark-specific
seed set, classifier, and Platt scaler are absent in this rule-based
setting, the embedding signal abstains and only the metadata signal
contributes in practice.

\subsection{Step-Level Cost Case Study: \texttt{sympy\_\_sympy-12096}}
\label{app:sympy}

Table~\ref{tab:appendix-sympy} shows a 13-step SWE-bench trajectory.
Plan~A is all-high execution with prompt caching; Plan~B is verified step-level routing
(the tier assignments in the ``Tier'' column).
The step-level routing achieves a 39.6\% cost reduction relative to all-high while resolving the issue.
Last-step output tokens are estimated~($*$).

\begin{table}[H]
\centering
\scriptsize
\caption{Step-level cost case study for \texttt{sympy\_\_sympy-12096}.
Plan~A = all-\texttt{high} with prompt caching.
Plan~B = verified step-level routing.
Token counts are reported as individual tokens; costs in USD.}
\label{tab:appendix-sympy}
\begin{adjustbox}{max width=\linewidth}
\begin{tabular}{r l r r r r r r}
\toprule
Step & Tier & Plan A in & Plan A out & Plan A cost & Plan B in & Plan B out & Plan B cost \\
\midrule
 1 & mid      & 1,321 & 112 & \$0.0111 & 1,284 & 108 & \$0.0005 \\
 2 & mid      & 1,744 &  71 & \$0.0051 & 1,699 &  68 & \$0.0003 \\
 3 & mid\_high & 1,846 &  69 & \$0.0032 & 1,647 &  66 & \$0.0003 \\
 4 & mid\_high & 2,502 &  67 & \$0.0067 & 2,343 &  65 & \$0.0003 \\
 5 & low      & 3,287 &  89 & \$0.0084 & 3,729 &  87 & \$0.0010 \\
 6 & mid\_high & 4,103 & 256 & \$0.0131 & 3,900 & 254 & \$0.0010 \\
 7 & low      & 4,512 &  55 & \$0.0060 & 5,215 &  55 & \$0.0009 \\
 8 & mid\_high & 4,612 & 168 & \$0.0071 & 4,403 & 168 & \$0.0007 \\
 9 & high     & 4,921 & 241 & \$0.0103 & 4,921 & 241 & \$0.0368 \\
10 & high     & 5,149 &  85 & \$0.0060 & 5,149 &  85 & \$0.0060 \\
11 & high     & 5,591 &  63 & \$0.0069 & 5,591 &  63 & \$0.0069 \\
12 & low      & 5,653 &  56 & \$0.0046 & 6,789 &  59 & \$0.0018 \\
13 & low      & 5,864 & 111$^{*}$ & \$0.0069 & 7,077 & 109$^{*}$ & \$0.0010 \\
\midrule
Total & -- & -- & -- & \textbf{\$0.0953} & -- & -- & \textbf{\$0.0576} \\
\bottomrule
\end{tabular}
\end{adjustbox}
\end{table}

\subsection{mtRAG/QMSum Judge Hardening Details}
\label{app:judge}

Table~\ref{tab:appendix-judge} documents the changes from the legacy judge rubric to the hardened version
used in the final dataset. Each row identifies a specific false-pass failure mode in the legacy judge
and the corresponding hardened rule applied in construction.

\begin{table}[H]
\centering
\small
\caption{mtRAG/QMSum judge hardening. The changes convert weak reference-similarity judging
into a conservative task-success predicate aligned with the current-turn/query requirement
and RadBench-style faithfulness criteria \citep{kuo2025radbench}.}
\label{tab:appendix-judge}
\begin{tabularx}{\linewidth}{p{0.22\linewidth}p{0.35\linewidth}X}
\toprule
Risk in legacy judge & Hardened rule & Purpose \\
\midrule
No primary criterion &
  First decide whether the answer completely and correctly resolves the current turn/query &
  Prevents answer/reference surface resemblance from substituting for task success \\
Reference treated as answer string &
  Treat reference as coverage hints; evidence/transcript is authoritative &
  Allows valid paraphrases and rejects reference-following errors \\
Unsupported facts &
  Hard fail for unsupported crucial numbers, dates, amounts, ratios, votes, or claims &
  Suppresses hallucinated but plausible-looking answers \\
Evidence contradiction &
  Hard fail when candidate contradicts retrieved passages or meeting transcript &
  Enforces faithfulness \\
Unanswerable/no-context cases &
  mtRAG pre-filter: must be answerable, have context passages, and positive human relevance feedback &
  Removes structurally unreliable samples before routing search \\
All tiers fail &
  Mark discarded; do not emit case JSON or supervision row &
  Prevents false passes from becoming verified labels \\
\bottomrule
\end{tabularx}
\end{table}

\subsection{Frontier LLM-as-Router Diagnostic Summary}
\label{app:opus}

Table~\ref{tab:opus-high} shows the Opus~4.6 predictions on verified-\texttt{high} SWE-bench steps;
Table~\ref{tab:appendix-opus-bias} provides additional diagnostics.
These numbers support the claim that a strong model's unexecuted routing judgment is not a reliable
substitute for execution-verified step-level supervision (\S\ref{sec:opus}).

\begin{table}[H]
\centering
\small
\begin{tabular}{lrrrrr}
\toprule
Verified tier & $\tilde{t}=\texttt{low}$ & $\tilde{t}=\texttt{mid}$ & $\tilde{t}=\texttt{mid\_high}$ & $\tilde{t}=\texttt{high}$ & Total \\
\midrule
$\hat{t}=\texttt{high}$ & 34 & 43 & 63 & \textbf{7} & 147 \\
\bottomrule
\end{tabular}
\caption{Opus 4.6 predictions on verified-\texttt{high} SWE-bench steps.
140 of 147 are routed below \texttt{high}.}
\label{tab:opus-high}
\end{table}

\begin{table}[H]
\centering
\small
\caption{Frontier LLM-as-router (Opus~4.6) diagnostic summary on SWE-bench.
The key finding is severe under-prediction of \texttt{high}-tier steps:
only 5\% of verified-\texttt{high} steps are predicted \texttt{high},
causing every SWE-bench trajectory to fail under this routing.}
\label{tab:appendix-opus-bias}
\begin{tabularx}{\linewidth}{lrr X}
\toprule
Diagnostic & Value & Scope & Interpretation \\
\midrule
SWE-bench valid rows & 300 & 336 rows total, 36 malformed & Valid subset for confusion-matrix analysis \\
Verified-high predicted high & 7/147 (4.8\%) & SWE-bench valid rows & Severe under-prediction of high-tier steps \\
Verified-high under-predicted & 140/147 (95.2\%) & SWE-bench valid rows & Most high steps routed below high \\
SWE-bench trajectory pass & 0/40 & Legacy Opus routing run & Every trajectory has $\ge$1 under-routed step \\
Middle-third under-prediction & 73\% & Step-position diagnostic & Core editing/test-running steps most under-predicted \\
First-third under-prediction & 46\% & Step-position diagnostic & Lower but still substantial under-routing \\
Last-third under-prediction & 53\% & Step-position diagnostic & Recovery/finalization steps remain difficult \\
\bottomrule
\end{tabularx}
\end{table}

\subsection{Label-Validation Audit Trail}
\label{app:validation}

Table~\ref{tab:appendix-validation} documents the validation mechanisms applied during dataset
construction. Together they ensure that no label enters the released JSONL unless the strong baseline
passed, a tier-pool model actually passes under the mixed-model trace, the judge does not mark
the answer as pseudo-passing, and the final human audit sees no surviving issue.

\begin{table}[H]
\centering
\small
\caption{Label-validation audit trail. Each component addresses a specific failure mode
in the construction pipeline.}
\label{tab:appendix-validation}
\begin{tabularx}{\linewidth}{p{0.22\linewidth}p{0.22\linewidth}p{0.28\linewidth}X}
\toprule
Validation component & Scope & Failure mode addressed & Release action \\
\midrule
Successful seed traces &
  All released instances &
  Conflating task failure with routing failure &
  Only passing seed traces enter degradation search \\
Sequential-locking downgrade search &
  Multi-step traces &
  Missing the last-passing tier; hypothetical overlays on the strongest trace &
  Push one tier below the Opus proposal under the causal mixed-model prefix until the harness fails, then lock the last passing tier \\
Tier cascade (3 models/tier) &
  All non-\texttt{low} tier labels &
  Single-model false negatives within a tier &
  Lock a cheaper tier only after an actual passing run \\
Strict mtRAG/QMSum judge &
  Open-ended RAG and summarization &
  False passes from weak judging &
  Use hard-fail rubric; discard all-tier failures \\
Final human audit &
  $\sim$10\% random step samples from each of BFCL, SWE-bench, and PinchBench &
  Still-downgradeable tier labels that the upstream search did not tighten to the conditional optimum &
  Human review under the router-visible prefix decides whether the released tier can be lowered by one more step; one SWE-bench step was flagged as further-downgradeable and its tier was lowered accordingly \\
\bottomrule
\end{tabularx}
\end{table}

\subsection{Manual Audit Protocol and Summary}
\label{app:manual-audit}

The manual audit is the final quality gate of the static track: it runs
\emph{after} the Opus-initialized downgrade search, tier-pool cascade check,
open-ended judge hardening, and prefix reconstruction of
\S\ref{sec:dataset-construction} are all complete. The audit is deliberately a
human-judgment pass rather than a second harness execution. Its purpose is to
cross-check that the released tier label is still defensible on the same
router-visible context that a deployed router would see, not to recompute any
pass predicate.

\paragraph{Sampling.}
We follow the GT tier-optimality review protocol included with the release
artifacts. The audit unit is a single routed step so that single-step and
multi-step cases are treated on the same granularity. Within each of the
three workloads that contain multi-step trajectories (BFCL, SWE-bench,
PinchBench) we draw an independent random sample of roughly 10\% of the
workload's routed steps. mtRAG and QMSum are single-turn tasks that lack
multi-turn compounding complexity; their labels are already constrained by the
hardened judge whose pseudo-pass patterns were eliminated through iterative
calibration (\S\ref{sec:dataset-construction}), and are not resampled in this
audit.

\paragraph{Review procedure.}
For each sampled step, the router-visible \texttt{messages} prefix is held
fixed at its released form. An optional large-model pre-label may be attached
as a reference, but the final decision is always taken by a human reviewer,
who asks whether lowering the verified tier by one more level would still
produce a response that correctly handles the current step; the step is then
labelled as \emph{tight}, \emph{uncertain}, or \emph{further-downgradeable}. A \emph{tight} verdict means the released tier
is already at the conditional optimum and cannot be lowered one more step
without breaking task resolution; a \emph{further-downgradeable} verdict means
the reviewer believes the released tier is still too high and the label can
be tightened by one more tier. The audit inspected 64 steps in total; one
SWE-bench step was flagged as further-downgradeable and its verified tier was
lowered by one tier before release, while the remaining 63 steps were judged
tight. This internal audit is intended to catch residual labels that the
upstream search did not tighten to the conditional optimum, and to clarify
label semantics; it is not presented as an external annotation study or as a
confidence interval for the full corpus.

\begin{table}[H]
\centering
\small
\caption{Manual audit summary. A sampled step is counted as
\emph{Downgradeable} when the reviewer believes its verified tier can still
be lowered by one more tier without breaking task resolution; \emph{Tight \%}
is the complementary share whose tier label is already at the conditional
optimum. Sampling is done at the step granularity across the three workloads
that contain multi-step trajectories; mtRAG and QMSum are single-turn tasks
already guarded by the hardened judge and are not resampled here. The audit is a targeted internal quality check.}
\label{tab:manual-audit}
\begin{tabular}{lrrrrr}
\toprule
Slice & Total steps & Sampled steps & Sampling \% & Downgradeable & Tight \% \\
\midrule
BFCL & 248 & 25 & 10.1 & 0 & 100.0 \\
SWE-bench & 336 & 34 & 10.1 & 1 & 97.1 \\
PinchBench & 48 & 5 & 10.4 & 0 & 100.0 \\
\midrule
Total & 632 & 64 & 10.1 & 1 & 98.4 \\
\bottomrule
\end{tabular}
\end{table}

\FloatBarrier
\noindent\begin{minipage}{\linewidth}
\paragraph{Examples of regenerated responses.}
We randomly sampled three cases from each of the five workloads and reviewed
one regenerated step per case, yielding 15 responses. All 15 responses were
judged locally sufficient. Table~\ref{tab:regenerated-examples} presents one
example per workload from this review. For each example, we held the released
prefix fixed and generated one response using a model from its labeled tier.
Among the displayed examples, GLM 4.5 Air generated the four low-tier responses,
and GPT-5.4 generated the high-tier SWE-bench response.
For these examples, GPT-5.4 reviewed the mtRAG, QMSum, and PinchBench outputs; Gemini 3.1 Pro
reviewed the BFCL and SWE-bench outputs. The review checked task relevance,
important omissions, tool selection, and arguments. For SWE-bench and
PinchBench, task-relevant tool schemas were reconstructed from the released
context and recorded calls. The displayed tool actions were inspected without execution
in the original task environments.\par
\end{minipage}\par

\begin{table}[H]
\centering\small
\begin{tabularx}{\linewidth}{p{0.12\linewidth}p{0.07\linewidth}XX}
\toprule
Workload & Tier & Regenerated response & Review \\
\midrule
BFCL & low & Calls both food and drink tools to remove anchovies from a salad
and update order 123. & Both requests covered; arguments match the tool schemas. \\
mtRAG & low & Explains how the cotton gin contributed indirectly to the Civil
War through the expansion of slavery. & Supported by the supplied passages. \\
QMSum & low & Summarizes remote-control functions, foldable design,
touchscreens, and charging options. & Main discussion covered, with minor
imprecision. \\
PinchBench & low & Proposes writing an approximately 480-word remote-work
blog to \texttt{blog\_post.md}. & Relevant content and appropriate file-write
arguments. \\
SWE-bench & high & Proposes a Python check of Sphinx's LaTeX highlighting
output. & Debugging action comparable to the recorded one; downstream effects
require execution. \\
\bottomrule
\end{tabularx}
\caption{One example per workload from the 15-case review (three cases per workload).
Reviews concern response content and proposed actions; new tool calls were
not executed in the original task environments.}
\label{tab:regenerated-examples}
\end{table}

\subsection{Fixed-Prefix Execution Repeatability}
\label{app:fixed-prefix-repeatability}

We fixed one published prefix per workload and evaluated an available
representative of its released tier. The SWE-bench case used high-tier
GPT-5.4; the BFCL, PinchBench, mtRAG, and QMSum cases were randomly sampled
from published mid-tier rows and used MiniMax M2.5. We report the initial
three repetitions for each prefix.

Target inputs, tool schemas, and generation parameters were held fixed.
Target generation used temperature~0.7, top-$p=1.0$, and a maximum output
length of 4,096 tokens, with no explicit random seed and application-level
response replay disabled. SWE-bench, BFCL, and PinchBench used their official
graders; mtRAG and QMSum used GPT-5.4 with the fixed structured rubric.
Multi-step outcomes include the effects of subsequent continuations.

\begin{table}[H]
\centering
\small
\begin{tabular}{lcr}
\toprule
Workload & Target tier & Successful executions \\
\midrule
SWE-bench & \texttt{high} & 3/3 \\
BFCL & \texttt{mid} & 2/3 \\
PinchBench & \texttt{mid} & 2/3 \\
mtRAG & \texttt{mid} & 2/3 \\
QMSum & \texttt{mid} & 2/3 \\
\bottomrule
\end{tabular}
\caption{Initial three repeated executions of one selected published prefix
per workload. Counts describe the selected cases and representative models
under the stated evaluation settings.}
\label{tab:fixed-prefix-repeatability}
\end{table}

All three SWE-bench executions passed the official tests; each other case
had one failed execution (Table~\ref{tab:fixed-prefix-repeatability}). Failures
involved selecting the wrong quantities from context (BFCL), omitting
calendar constraints (PinchBench), and making unsupported assertions
(mtRAG and QMSum). Execution outcomes therefore varied despite fixed target
inputs and generation parameters.

Our labels are conditional on the realized router-visible prefix, rather
than on the task alone. Different successful trajectories may therefore
yield different prefixes and sufficient tiers; agreement across such
trajectories is not required by our formulation, which targets per-call
sufficiency rather than a globally optimal routing path. This variation can
also enrich the static corpus with diverse routing contexts, provided that
each new prefix--label pair is independently verified. Repeated execution
of an identical prefix may still vary, so a released label denotes
sufficiency verified under our construction protocol, not guaranteed
success on every run.

\subsection{Static-Track Breakdown}
\label{app:static-breakdown}

\paragraph{Per-workload cost savings.}

Table~\ref{tab:per-benchmark} shows that SWE-bench is the hardest static slice.
It carries 34.64\% of the row-weighted \textsc{CostSave} score, and every
router has negative cost savings there because one under-routed step fails the
whole trajectory and triggers the failure-aware bill. On the other four
workloads, SR-KNN and ClawRouter (rule-based) both exceed 85\% cost savings on average; the
global ranking depends on whether a router preserves code-repair trajectories.

\begin{table}[H]
\centering
\scriptsize
\begin{tabular}{lrrrrrr}
\toprule
Router & BFCL & mtRAG & QMSum & Pinch & SWE-bench & Overall \\
\midrule
SR-KNN       & \textbf{90.49} & \textbf{92.35} & 90.24 & 35.36 & \textbf{-1.64} & \textbf{56.18} \\
ClawRouter (rule-based) & 89.87 & 83.45 & \textbf{92.12} & \textbf{55.20} & -6.55 & 53.82 \\
UncommonRoute (rule-based) & 47.14 & 91.93 & 90.24 & 39.80 & -86.08 & 15.99 \\
\midrule
Row weight  & 25.57 & 19.90 & 14.95 & 4.95 & 34.64 & 100.00 \\
\bottomrule
\end{tabular}
\caption{Per-workload \textsc{CostSave} under the static scoring protocol.
The SWE-bench slice is the main stress test: failed trajectories trigger the
failure-aware numerator in \S\ref{sec:static-scoring}.}
\label{tab:per-benchmark}
\end{table}

\paragraph{Router failure modes.}
\label{sec:error-patterns}

The three baselines fail in different ways (Table~\ref{tab:error-patterns}). On
the 800 rows whose verified tier is below \texttt{high}, SR-KNN is mostly
exact; ClawRouter (rule-based) almost always routes one tier above the label; UncommonRoute
(rule-based) is exact more often than ClawRouter (rule-based) but jumps to \texttt{high}
217 times.

\begin{table}[H]
\centering
\small
\begin{tabular}{lrrr}
\toprule
Prediction pattern on $\hat{t}<\texttt{high}$ & SR-KNN & ClawRouter (rule-based) & UncommonRoute (rule-based) \\
\midrule
Exact tier                         & \textbf{627} & 107 & 488 \\
Up one tier                         & 10 & \textbf{649} & 47 \\
Jump to \texttt{high}               & 117 & 21 & \textbf{217} \\
Fail: predicted below $\hat{t}$     & 46 & 23 & \textbf{48} \\
\bottomrule
\end{tabular}
\caption{Error patterns on the 800 non-\texttt{high} rows.}
\label{tab:error-patterns}
\end{table}

SWE-bench exposes why row-level behavior is insufficient
(Table~\ref{tab:swe-high}). UncommonRoute (rule-based) recognizes many
verified-\texttt{high} steps, but missing even one step in an 8--13 call
trajectory fails the instance.

\begin{table}[H]
\centering
\small
\begin{tabular}{lrrr}
\toprule
Router & Predicted \texttt{high} on $\hat{t}=\texttt{high}$ & Hit rate & Failed traj. \\
\midrule
SR-KNN       & 137/168 & \textbf{81.5\%} & \textbf{10/40} \\
ClawRouter (rule-based) & 7/168   & 4.2\%  & 38/40 \\
UncommonRoute (rule-based) & 105/168 & 62.5\% & 39/40 \\
\bottomrule
\end{tabular}
\caption{SWE-bench high-tier decisions. One under-routed critical step fails
the whole trajectory.}
\label{tab:swe-high}
\end{table}

\subsection{Cost-Savings Accounting Ablation}
\label{app:cost-ablation}

This section provides the full derivation and per-variant numbers referenced in \S\ref{sec:main-results}.

\paragraph{Pricing asymmetry drives the ablation.}
Table~\ref{tab:save-asymmetry} reports, for three representative step profiles, the per-step saving against the always-\texttt{high} baseline under each target tier. The ratio \texttt{save\_mid / save\_low} is within 3\% of $1$ across all three step sizes, which means a router that mis-routes from $\hat{t}=\texttt{low}$ to $\tilde{t}=\texttt{mid}$ loses almost no money in absolute terms.

\begin{table}[H]
\centering
\small
\caption{Savings vs.\ always-\texttt{high} for different target tiers on three representative step profiles.
Mis-routing from \texttt{low} to \texttt{mid} is nearly cost-free.}
\label{tab:save-asymmetry}
\begin{tabular}{lrrrr}
\toprule
Step profile & save(high$-$low) & save(high$-$mid) & save(high$-$mid\_high) & save\_mid / save\_low \\
\midrule
4k cold step     & 3.621¢ & 3.530¢ & 3.467¢ & 0.975 \\
20k warm step    & 1.965¢ & 2.032¢ & 1.900¢ & 1.034 \\
100k cold step   & 61.13¢ & 60.65¢ & 62.67¢ & 0.992 \\
\bottomrule
\end{tabular}
\end{table}

\paragraph{Five scoring variants.}
We compare five $N/D$ accounting schemes on the same predictions:

\begin{itemize}
\item \textbf{A} (legacy row-level denominator): $D = \sum_i (\mathrm{cost}(3;i) - \mathrm{cost}(\hat{t}_i;i))$ over passing rows with $\hat{t}_i \ne \texttt{high}$; $N$ ignores failures.
\item \textbf{B} (legacy all-row denominator): $D$ as in A but over all rows; failing trajectories subtract $\sum_i \mathrm{cost}(\tilde{t}_i;i)$ at trajectory level.
\item \textbf{C} (intermediate high-row exclusion): exclude $\hat{t}_i=\texttt{high}$ rows from $D$.
\item \textbf{D} ($D = \sum_i \mathrm{cost}(3;i)$, failed-step penalty at step level, no trajectory-level penalty).
\item \textbf{E} (final failure-aware formula): same $D$ as D, but failed trajectories charge $-\sum_i \mathrm{cost}(\tilde{t}_i;i)$ at trajectory level.
\end{itemize}

\begin{table}[H]
\centering
\small
\caption{Ablation of \textsc{CostSave} accounting. Variants A--D all rank ClawRouter (rule-based) above SR-KNN;
only variant E (the final failure-aware formula) aligns with \textsc{RowPass}, \textsc{RowExact}, and \textsc{TrajPass}.}
\label{tab:ablation-scoring}
\begin{tabular}{lrrr}
\toprule
Variant & SR-KNN \textsc{cost}\% & ClawRouter (rule-based) \textsc{cost}\% & Ranking \\
\midrule
A (legacy row-level denominator)                  & 84.99 & \textbf{89.71} & Claw wins \\
B (legacy all-row denominator)                    & 84.45 & \textbf{84.59} & Claw wins \\
C (intermediate high-row exclusion)               & 79.11 & \textbf{91.31} & Claw wins \\
D ($D=\sum\mathrm{cost}(3)$, step-level fail)    & 61.81 & \textbf{67.67} & Claw wins \\
\textbf{E} (final failure-aware formula)          & \textbf{56.18} & 53.82 & SR wins \\
\bottomrule
\end{tabular}
\end{table}

\paragraph{Why variant E recovers the consistent ranking.}
Variant E interprets a failing trajectory as ``the router burned its own (cheap) budget \emph{and} the user paid the full \texttt{high} bill to recompute.'' Formally, the user-side bill on a failing trajectory is $\sum_i \mathrm{cost}(\tilde{t}_i;i) + \sum_i \mathrm{cost}(3;i)$, so savings against always-\texttt{high} equal $-\sum_i \mathrm{cost}(\tilde{t}_i;i)$, which is exactly what $N$ encodes. The $\sum_i \mathrm{cost}(3;i)$ retry term is absorbed by $D$ rather than double-counted, keeping \textsc{CostSave} $\in [-\infty, 100]$ with $0$ representing always-\texttt{high} routing. On the full benchmark, ClawRouter (rule-based) has 649 up-one errors (near-free in absolute terms) that can no longer compensate for its 38 failing SWE-bench trajectories, and its \textsc{CostSave} aligns with its \textsc{RowPass}/\textsc{RowExact}/\textsc{TrajPass} figures.

\subsection{Opus 4.6 Confusion Matrix}
\label{app:opus-confusion}

For reference, the Opus 4.6 routing run recorded under an earlier three-metric accounting protocol reports an overall of 84.05, with row pass $=$85.77, row exact $=$82.88, and cost savings $=$83.48. This is not directly comparable to the static scores in Table~\ref{tab:main-results} because the earlier accounting does not apply the failure-aware trajectory penalty in \S\ref{sec:static-scoring} and does not report a trajectory-pass metric; we therefore treat it as a historical data point rather than a ranked baseline.

Table~\ref{tab:opus-confusion} shows the full confusion matrix summarized in \S\ref{sec:opus}.

\begin{table}[H]
\centering
\small
\caption{Opus 4.6 routing predictions vs.\ verified tiers on the 300 valid SWE-bench rows
(36 malformed outputs excluded). Diagonal cells are highlighted. Opus rarely commits to
\texttt{high}, even when the verified tier is \texttt{high}.}
\label{tab:opus-confusion}
\begin{tabular}{lrrrrr}
\toprule
 & $\tilde{t}=\texttt{low}$ & $\tilde{t}=\texttt{mid}$ & $\tilde{t}=\texttt{mid\_high}$ & $\tilde{t}=\texttt{high}$ & row total \\
\midrule
$\hat{t}=\texttt{low}$       & \textbf{52 (61\%)} & 15 & 17 & 1 & 85 \\
$\hat{t}=\texttt{mid}$       & 13 & \textbf{7 (23\%)} & 10 & 0 & 30 \\
$\hat{t}=\texttt{mid\_high}$ & 8  & 13 & \textbf{16 (42\%)} & 1 & 38 \\
$\hat{t}=\texttt{high}$      & 34 & 43 & 63 & \textbf{7 (5\%)} & 147 \\
\bottomrule
\end{tabular}
\end{table}

\subsection{mtRAG Row Example}
\label{app:mtrag-example}

A second row format, complementing the SWE-bench example in \S\ref{sec:benchmark-overview}, illustrates how \texttt{low}-tier labels arise: after several retrieval turns, a short follow-up query can be answered by the cheapest tier in the pool.

\begin{center}
\begin{minipage}{0.95\linewidth}
\scriptsize
\begin{verbatim}
{
  "id": "mtrag_..._turn_3_step_1",
  "benchmark": "mtrag",
  "step_index": 1, "total_steps": 1,
  "messages": [
    {"role": "system", "content": "Answer based on the retrieved passages..."},
    {"role": "user",   "content": "[Passage] Major.minor update... [Passage] ..."},
    {"role": "assistant", "content": "..."},
    ...multi-turn history...
    {"role": "user",   "content": "Major.minor update."}
  ],
  "target_tier": "low", "target_tier_id": 0
}
\end{verbatim}
\end{minipage}
\end{center}

\end{document}